\documentclass[lettersize,journal]{IEEEtran}
\usepackage{amsmath,amsfonts}
\usepackage{algorithmic}
\usepackage{algorithm}
\usepackage{array}
\usepackage[caption=false,font=normalsize,labelfont=sf,textfont=sf]{subfig}
\usepackage{textcomp}
\usepackage{stfloats}
\usepackage{url}
\usepackage{verbatim}
\usepackage{graphicx}
\usepackage{cite}
\usepackage{orcidlink}
\begin{document}

\title{A Soft Robotic Interface \\for Chick-Robot Affective Interactions}

\author{Jue Chen$^{\orcidlink{0009-0001-9138-6927}}$~\IEEEmembership{Student Member,~IEEE}, Alexander Mielke$^{\orcidlink{0000-0002-8847-6665}}$, Kaspar Althoefer$^{\orcidlink{0000-0002-1141-9996}}$ ~\IEEEmembership{Fellow,~IEEE}, and Elisabetta Versace
\thanks{This work involved animals in its research. Approval of experimental procedures was granted by the Queen Mary University of London ethics committee (AWERB) and Home Office (PP5180959).}
\thanks{During the preparation of this work, the authors used \textit{ChatGPT} 5.4 to improve the readability of the paper.}
\thanks{This work of J. Chen was supported by the China Scholarship Council (CSC, No. 202408530124). The work of E. Versace was supported in part by the Leverhulme Trust research grant RPG-2020-287 and in part by the Royal Society Leverhulme Trust fellowship SRF\textbackslash R1\textbackslash 21000155.}
\thanks{J. Chen and K. Althoefer are with the Centre for Advanced Robotics @ Queen Mary,
School of Engineering and Materials Science, Queen Mary University of London, United Kingdom. (e-mail: jue.chen@qmul.ac.uk).}
\thanks{A. Mielke and E. Versace are with the School of Biological and Behavioural Sciences, Queen Mary University of London, United Kingdom.}
}



\maketitle

\begin{abstract}
The potential of Animal-Robot Interaction (ARI) in welfare applications depends on how much an animal perceives a robotic agent as socially relevant, non-threatening and potentially attractive (acceptance). Here, we present an animal-centered soft robotic affective interface for newly hatched chicks (\textit{Gallus gallus}). The soft interface provides safe and controllable cues, including warmth, breathing-like rhythmic deformation, and face-like visual stimuli. We evaluated chick acceptance of the interface and chick-robot interactions by measuring spontaneous approach and touch responses during video tracking. Overall, chicks approached and spent increasing time on or near the interface, demonstrating acceptance of the device. Across different layouts, chicks showed strong preference for warm thermal stimulation, which increased over time. Face-like visual cues elicited a swift and stable preference, speeding up the initial approach to the tactile interface. Although the breathing cue did not elicit any preference, neither did it trigger avoidance, paving the way for further exploration. These findings translate affective interface concepts to ARI, demonstrating that appropriate soft, thermal and visual stimuli can sustain early chick-robot interactions. This work establishes a reliable evaluation protocol and a safe baseline for designing multimodal robotic devices for animal welfare and neuroscientific research.
\end{abstract}


\section{Introduction}
\IEEEPARstart{A}{nimal-Robot} Interaction (ARI), as an extension of Human-Robot Interaction (HRI), has growing potential for animal welfare. For instance, robotic devices can provide environmental enrichment through diverse and scalable interactions\cite{adamsDesignDevelopmentAutonomous2021,kimAnimalRobotInteractionPet2009}. Robots can also serve as interactive partners, helping to understand perception, learning and social responses in animals\cite{schneidersDesigningMultispeciesWorlds2024,sloninaUsingRoboChickIdentify2021,gribovskiyMixedSocietiesChickens2010,demargerieInfluenceMobileRobot2011,jollyAnimaltorobotSocialAttachment2016,vaughanExperimentsAutomaticFlock2000}. However, the benefits of ARI depend on the ability to create devices that are 'accepted' by animals. Acceptance is the degree to which an animal perceives a robotic agent as a socially relevant, non-threatening and potentially attractive component of its environment. This differs from task completion, such as moving a flock toward a goal \cite{vaughanExperimentsAutomaticFlock2000}. If an animal does not perceive a robotic device as safe and meaningful, observed behaviors may reflect stress or mere curiosity  \cite{costaAspectsChickenBehavior2012}. Animal acceptance is therefore a crucial metric to assess whether a robotic device is not only safe but also contextually appropriate.

In HRI, affective touch has been widely studied  \cite{yohananRoleAffectiveTouch2012,flaggAffectiveTouchGesture2013,haynesCalmingHugDesign2022,onishiMoffulyIIRobotThat2024,tsirkaTouchHumanSocial2024}. One emerging trend is delivering affective touch through soft robotic interfaces \cite{haynesCalmingHugDesign2022,sabinsonEveryBreathTesting2024,liuBreatHapticsEnablingGranular2024}. Their inherent compliance enables the safe delivery of comforting stimuli, such as gentle contact, warmth, and rhythmic deformation \cite{haynesCalmingHugDesign2022,sabinsonEveryBreathTesting2024,liuBreatHapticsEnablingGranular2024,schirmerEditorialOverviewAffective2022}. Here we use the approach on newly-hatched domestic chicks (\textit{Gallus gallus}), a species with a wide socioeconomic impact, involved in an expansion of robotic applications in farming \cite{ozenturkRoboticsPoultryFarming2024a}. 

\begin{figure}
\centering
\includegraphics[width=3.4in]{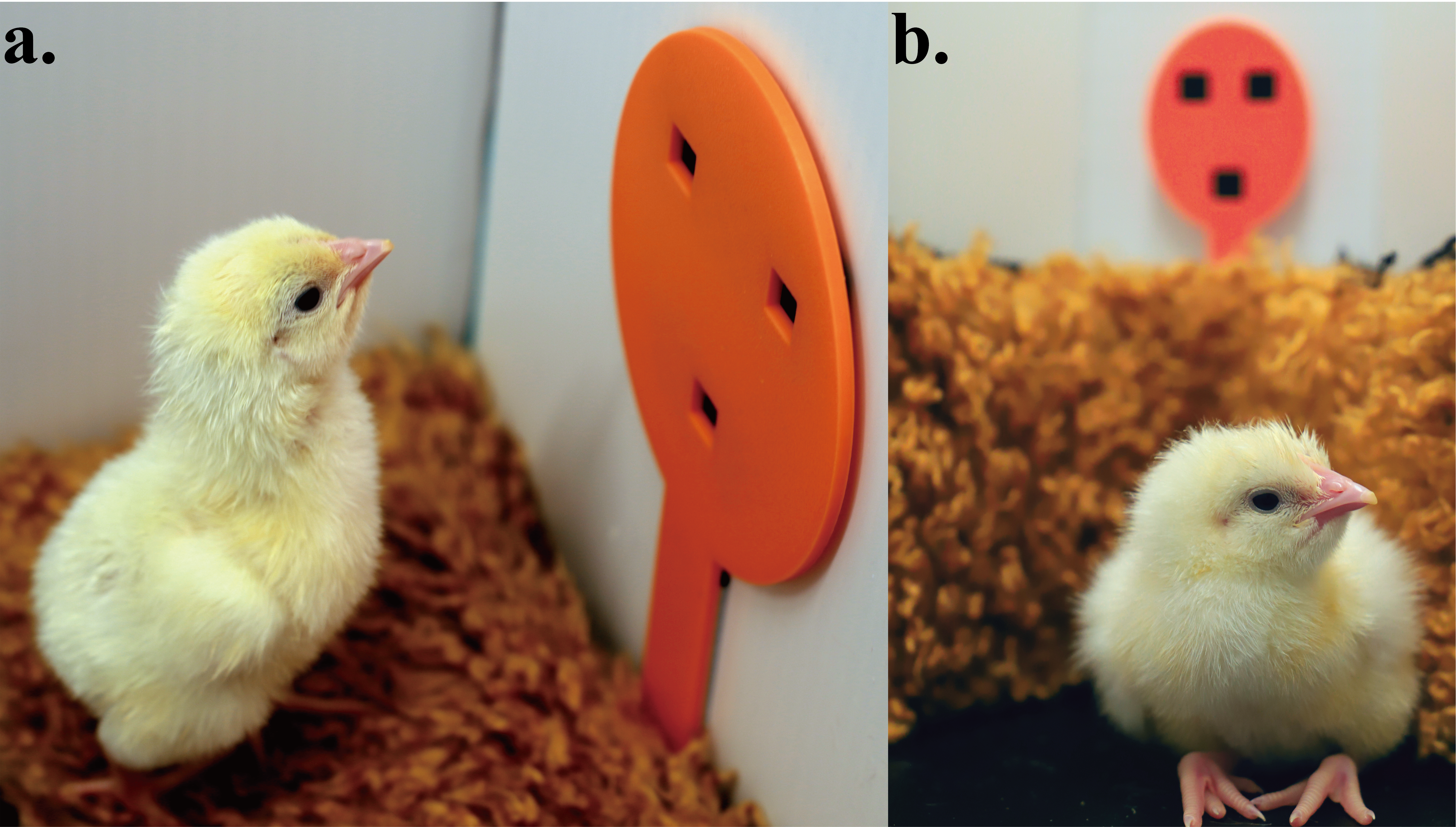}
\caption{Examples of chick–soft robot affective interface interactions in two experiments. \textbf{a.} Experiment 2 (horizontal, with faceplate and heating, see \ref{fig_2}a): chick looking toward the faceplate visual cue. \textbf{b.} Experiment 3 (vertical, with faceplate and heating, see \ref{fig_2}b): chick maintaining continuous contact with the heated interface.}
\label{fig_1}
\end{figure}

The few ARI studies conducted with chicks (e.g.,\cite{gribovskiyMixedSocietiesChickens2010,demargerieInfluenceMobileRobot2011,jollyAnimaltorobotSocialAttachment2016,sloninaUsingRoboChickIdentify2021}) have made clear the advantages of this model. First, chicks are precocial and attracted to moving objects from the first hours after hatching \cite{versaceMultipleWeakBiases2026}, enabling early interactions, controlled tests for spontaneous attraction and easy handling \cite{versaceOriginsKnowledgeInsights2015,rosa-salvaSensitivePeriodsSocial2021a}. Moreover, newly hatched chicks exhibit filial imprinting, a fast learning mechanism based on simple exposure, which in a few minutes produces a strong social attraction toward the first conspicuous objects that chicks experience after hatching \cite{bolhuisMechanismsAvianImprinting1991,jollyAnimaltorobotSocialAttachment2016,mccabeVisualImprintingBirds2019,versaceFirstsightRecognitionTouched2024}. Importantly for robotics, chicks can quickly imprint on artificial objects, including those that are non-naturalistic \cite{wangSpontaneousBiasesEnhance2024,freelandAssessingPreferencesAdult2025}, thus simplifying hardware development. The initial spontaneous attraction to moving objects \cite{rosa-salvaSensitivePeriodsSocial2021a,versaceMultipleWeakBiases2026} and subsequent imprinting generate characteristic affiliative behaviors that can be quantified by the time spent near the stimulus or within its immediate surrounding area \cite{sloninaUsingRoboChickIdentify2021,wangSpontaneousBiasesEnhance2024,freelandAssessingPreferencesAdult2025}. These measures are drawn from spontaneous behaviors and require minimal intervention, making them a direct and welfare-friendly index of attraction and preference\cite{sloninaUsingRoboChickIdentify2021,wangSpontaneousBiasesEnhance2024,versaceFirstsightRecognitionTouched2024,freelandAssessingPreferencesAdult2025}, that can be used as a proxy for acceptance. We use this approach based on the strong social motivation that chicks exhibit at the beginning of life (conversely, they spontaneously avoid potentially threatening stimuli \cite{hebertInexperiencedPreysKnow2019}). 

In this study, we bridge HRI and ARI by developing a soft robotic affective interface for newly hatched chicks. Moving from human-centered  to animal-centered designs, we explicitly incorporate cues relevant for the early stages of chick life: soft interface materials, warmth, soft robotic rhythmic breathing, and face-like visual features. The primary contribution of this paper is the design and systematic validation of an ARI interface integrating multiple affective cues. Across four experiments, we isolate specific affective cues and manipulate spatial arrangements of the interface to measure their impact on chick-robot interactions. We demonstrate that thermal and visual cues effectively drive immediate and sustained preferences, while soft robotic breathing is behaviorally neutral, triggering neither preference nor avoidance. This work provides a validated robotic platform and identifies clear design priorities for future animal-welfare applications.

\begin{figure*}
\centering
\includegraphics[width=7in]{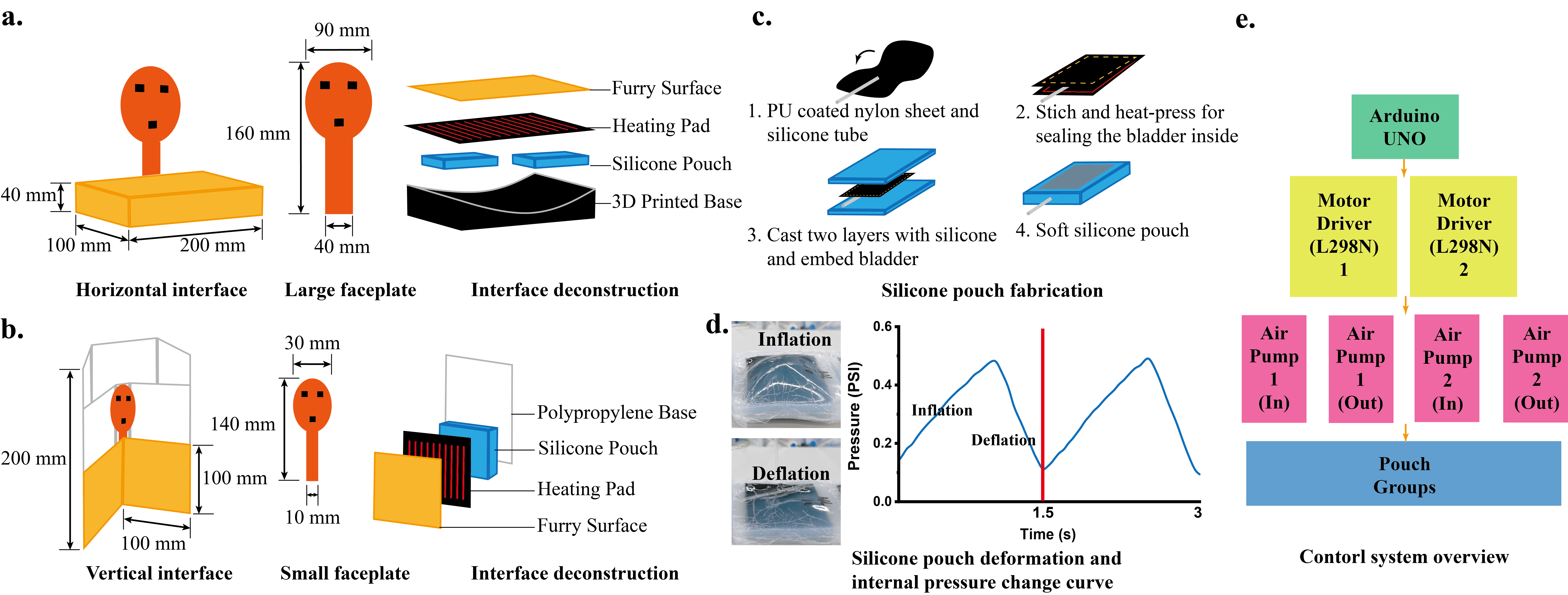}
\hfil
\caption{System design. \textbf{a.} Horizontal interface with a large faceplate. \textbf{b.} Vertical interface with a small faceplate. Both interfaces share the same structure: soft furry surface, heating pad, silicone pouch and a base (3D-printed in the horizontal interface; polypropylene in the vertical interface). \textbf{c.} The key steps in silicone pouch fabrication (embedded bladder, sealing, and two-layer silicone casting). \textbf{d.} Pouch surface deformation and internal pressure profile during one actuation cycle. \textbf{e.} Control architecture: an Arduino Uno controls two L298N motor drivers that switch between four air pumps (air in or out for two pouch groups).}
\label{fig_2}
\end{figure*}

\section{System Design}
\subsection{Overall Requirements}
Working within an animal-centered ethical framework \cite{manciniAnimalcentredEthicsAnimal2017a}, we adopt a controllable, low-risk, and withdrawable interaction strategy. The interface was designed to support animal welfare and to match the physiological characteristics and natural preferences of newly hatched chicks. The interface was physically safe, with stimulus intensities set within the chicks' comfort range. First, the size of the interface was scaled to the animals \cite{versaceOriginsKnowledgeInsights2015,rosa-salvaSensitivePeriodsSocial2021a,versaceMultipleWeakBiases2026}, so that it was easy to perceive without appearing threatening. Second, the interactive cues were designed and adjusted based on known chick preferences, including spontaneous attraction to warmth and face-like visual patterns \cite{deatonEffectBroodingTemperature1996,rosa-salvaFacesAreSpecial2010a,kobylkov_face_2024}. Third, the experimental arena provided enough open space for chicks to explore freely and move away from the interface at any time. Beyond this, we had planned to stop trials if signs of distress were observed, although this proved unnecessary.

\subsection{Soft Robotics Affective Interface Design}
We designed two soft robot affective interface prototypes, one horizontal and one vertical (Fig. \ref{fig_2}a--b). Both shared the same soft contact furry surface, heating pad and silicone pouch layers but differed in the size of face-like visual cues and the base layer constituents.

\textbf{Soft furry surface:} The soft furry surface - the principal contact area - was made of a fur-like, honey-colored material. It provided a soft, uniform contact area, helped create a comfortable interaction environment, and separated the chick from the internal components. This cover followed the outer shape of the device and fully enclosed the underlying structure. It was secured by Velcro attachments, allowing easy removal for cleaning and replacement.

\textbf{Heating Pad:} The heating pad layer provided the main thermal cue of the interface and consisted of a thin, flexible carbon-fiber pad (Fig. \ref{fig_2}a--b). It was placed directly beneath the furry surface to provide controlled warmth above ambient temperature. The pad ($85 \times 65$ mm) was held in a fabric pocket that fixed its position, reduced movement under repeated loading, prevented local folding, guided the wire path, and provided strain relief. It was powered at 5 V / 5 A, with a rating of 7 W. For animal welfare and protection, the maximum surface temperature was limited to 35 °C \cite{deatonEffectBroodingTemperature1996}. When activated, the heating layer maintained a surface temperature of $33 \pm 1$ \textdegree C, providing a stable increase above room temperature ($28.5 \pm 1$ \textdegree C) while remaining within the safe range.

\textbf{Silicone Pouch:} This layer provided the dynamic function of the interface and consisted of two side-by-side silicone pouches cast from Ecoflex 10 (Fig. \ref{fig_2}c). It generated a gentle, breathing-like rhythmic deformation on the contact surface. Each pouch ($80 \times 80$ mm) had a thickness of 5 mm and was driven by 12 V DC air pumps. Based on the reported chick respiratory rate range (30--90 cycles/min, 0.5--1.5 Hz) \cite{moriyaSimultaneousMeasurementsInstantaneous2003}, we set the actuation cycle period to 1.5 s, which lies within the physiological range and represents an intermediate breathing rate. During actuation, internal pouch pressure was regulated between 0.1 and 0.5 psi (Fig. \ref{fig_2}d) to maintain safe and stable operation.
 
\textbf{Faceplate:} The faceplate served as the main visual cue of the interface and was 3D printed in orange (Fig. \ref{fig_2}a--b). It provided a clear visual target and included three square openings arranged in a face-like pattern, representing two “eyes” and one “mouth”. Face-like configurations have been shown to attract newly hatched chicks \cite{rosa-salvaFacesAreSpecial2010a,kobylkov_face_2024}, humans \cite{valenza_face_1996}, and other animals \cite{versaceEarlyPreferenceFacelike2020}. To fit the two prototype layouts, the faceplate was produced in two sizes. The large version, used for the horizontal interface, had an oval head 90 mm wide, a narrow neck extension 30 mm wide, and an overall height of 160 mm. The small version, used for the vertical interface, had an oval head 40 mm wide, a neck extension 10 mm wide, and an overall height of 145 mm.

\begin{figure*}
\centering
\includegraphics[width=6in]{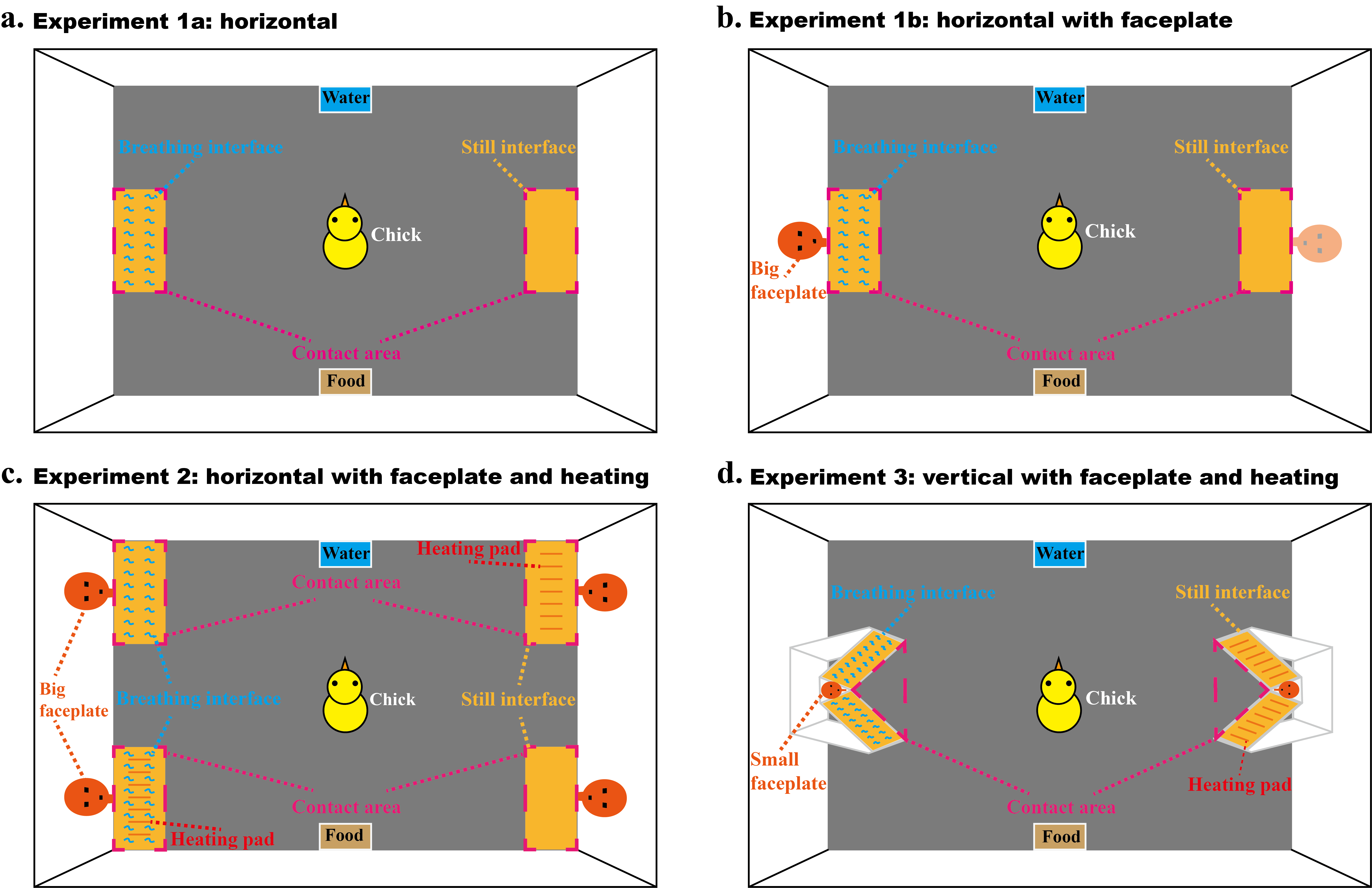}
\hfil
\caption{Experimental setups. \textbf {a–b.} Experiments 1a and 1b: two non-heated horizontal interfaces were placed at the midpoints of the short sides of the arena. In Experiment 1b, one big faceplate was attached to either the left or right interface, counterbalanced across sessions. \textbf{c.} Experiment 2: four horizontal interfaces were placed at the four corners, with two diagonally opposite interfaces heated and fixed. All interfaces were paired with big faceplates. \textbf{d.} Experiment 3: two vertical interfaces were placed at the midpoints of the short sides; the right interface was heated and fixed, and both interfaces were paired with small faceplates. Heated components are shown in red lines, and pink dashed regions mark the contact areas (where used to compute preferences). Water and food locations are indicated. All interfaces are honey-colored. During each session, just one side was in the "breathing state" (as indicated by the wavy blue lines). The two sides alternated every five minutes across the session.}
\label{fig_3}
\end{figure*}

\textbf{Base Layer:} The base layer served as the structural support of the interface and was made of 3D-printed parts (Fig. \ref{fig_2}a) and polypropylene (Fig. \ref{fig_2}b). It defined the overall geometry of the device, supporting the pouch assembly, and protecting the internal components. The horizontal base ($200 \times 100$ mm) has a nest-like shape, with raised edges and a slightly recessed center. For the vertical prototype, the base consists of a polypropylene support structure. The structure is 200 mm high, and the two pouch carrying panels are joined at a 90° angle. One pouch is mounted on each panel, forming a wing-like configuration. The base also provides repeatable boundary conditions for the pouches and routes pneumatic tubing and electrical connections to reduce external clutter and improve robustness.

\subsection{Hardware and Control}
An Arduino Uno (ATmega328P) served as the main controller (Fig. \ref{fig_2}e). Commands were received by an IR receiver module (KY-022, 38 kHz demodulation) and decoded on board to trigger predefined actuation routines. A remote controller reduced disturbance from human presence during experiments. The controller output digital signals to two L298N dual H-bridge driver boards, which independently controlled four DC pumps. The actuation module used four 12 V DC air pumps arranged as two functional pairs, with one pair assigned to each pouch group. Each pair included an inflation pump (air in) and a deflation pump (air out). This configuration supported group-wise pressure modulation and rapid releasing for safety and state reset. Pumps were connected to the silicone pouches via PVC tubing (4 mm inner diameter). To avoid noise disturbance, the pump assembly was placed inside an enclosure with soundproof foam and positioned away from the experimental arena. For stable operation, pump power was isolated from the logic supply. The Arduino was powered from a regulated 5 V source, while the pumps were driven by a dedicated 12 V source. For safety, the control logic enforces full deflation before any group switches to avoid pressure accumulation and to maintain a consistent initial state. In addition, a maximum pump-on time limit was imposed as a fail-safe.

\section{Experiment Design}
\subsection{Subjects and rearing conditions}
The experimental procedures involving animals were approved by the Queen Mary University of London ethics committee (AWERB) and Home Office (PP5180959).
We tested 61 newly hatched domestic chicks (\textit{Gallus gallus}) of the Ross 309 strain, in the first 24 hours after hatching: 15 chicks (6 females, 9 males) in Experiment 1a, 10 chicks (5 females, 5 males) in Experiment 1b, 22 chicks (12 females, 10 males) in Experiment 2 and 14 chicks (5 females, 9 males) in Experiment 3. The eggs were ordered from a qualified supplier (PD Hook, UK) and were incubated for 21 days under standard controlled conditions in darkness (37.7 \textdegree C and 40--60 \% humidity). Room temperature was controlled at $29 \pm 1$ \textdegree C.

\subsection{Apparatus}
Experiments were conducted in a wooden arena ($900 \times 600 \times 600$ mm), covered with a white polypropylene sheet inside and a non-slip black mat on the floor. A food container was placed at the midpoint of one of the long sides of the arena, and a water container was placed facing it on the opposite side. A high-definition video camera (Logitech C920S Pro Webcam) was positioned above the arena and recorded the experiments at 10 frames per second, at a resolution of $1280 \times 720$ pixels.

In Experiments 1a and 1b, two soft robotic affective interfaces (without heating) were placed horizontally at the midpoints of the short sides of the arena (Fig. \ref{fig_3}a--b). In Experiment 2, four soft-robotic interfaces were placed horizontally at the four corners (Fig. \ref{fig_3}c); two of which were heated and positioned diagonally across from each other.  In Experiment 3, two soft robotic interfaces were mounted vertically at the midpoints of the short sides; one of these was heated and positioned on the right hand side (Fig. \ref{fig_3}d).

Face-like visual cues were included in three experiments (Fig. \ref{fig_3}b--d). In Experiment 1b, a single large faceplate was fitted to one interface, on either the left or right side; this was counterbalanced across subjects. In Experiment 2, four big faceplates were fitted to the four interfaces. In Experiment 3, two small faceplates were fitted to the two interfaces.

\subsection{Procedure}
The same procedure was followed in each experiment. Each session lasted 30 minutes. A healthy chick with no prior visual exposure to conspecifics was placed at the center of the arena and allowed to freely explore the arena. Sessions were aborted and excluded from analysis if the chick showed no movement within the first 10 min.

\begin{figure*}
\centering
\includegraphics[width=7in]{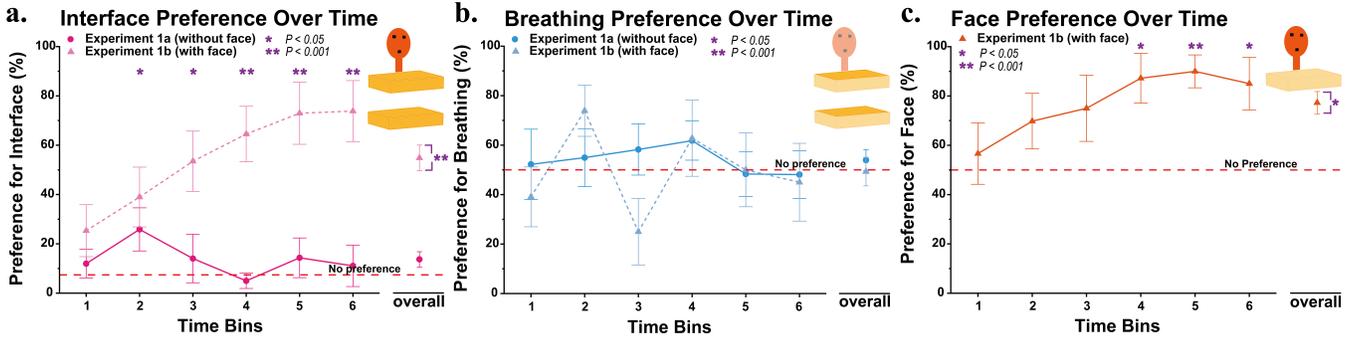}
\hfil
\caption{Preferences in Experiment 1a (horizontal) and Experiment 1b (horizontal, with face). Figures display raw data for visualization, whereas all data statistics were based on the fitted Beta mixed-effects models. The x-axis represents the time of the experimental session, divided into six consecutive 5-minute time bins. The isolated “overall” data points on the right of each panel indicate the estimated marginal means averaged across all six bins for the entire session. Red dashed lines indicate the chance levels. Error bars represent the standard error of the mean (SEM). Asterisks denote statistical significance (*$p < 0.05$, **$p < 0.001$). \textbf{a.} Interface preference over time. Experiment 1a (solid line) and Experiment 1b (dashed line). \textbf{b.} Breathing preference over time. Experiment 1a (solid line) and Experiment 1b (dashed line). \textbf{c.} Face preference over time in Experiment 1b.}
\label{fig_4}
\end{figure*}

During each session, rhythmic breathing from the silicone pouches was applied to only one side at a time. Each side was actuated for 5 min, alternating throughout the 30-minute session. The starting side was counterbalanced across subjects to ensure equal exposure to left and right actuation. The interface with thermal stimulation was maintained in the same position throughout the session.

\subsection{Data Analysis}
Chicks’ preference for the soft robotic affective interface was evaluated from recorded videos. A total of 54 valid recordings were included in the final analysis (10 in Experiment 1a, 10 in Experiment 1b, 20 in Experiment 2 and 14 in Experiment 3). Across the sub-experiments, 7 subjects were excluded: 5 in Experiment 1a (4 for no approach response and 1 for a recording failure) and 2 in Experiment 2 (both for no approach response).

To ensure reliable behavioral tracking, separate DeepLabCut models \cite{mathisDeepLabCutMarkerlessPose2018} were trained for each experimental setup. The likelihood threshold was determined empirically based on preliminary inspection of the tracking outputs. Recordings were retained only if at least 90\% of frames had a tracking likelihood of 0.6 or higher. A custom Python script then selected the highest-likelihood frame within each one-second interval, and short gaps caused by consecutive low-confidence frames were filled by linear interpolation.

Using the positional relationship between the chick and the interface coordinates, we defined a preference for the soft robotic affective interface using the following formulas (where \textit{P} denotes preference and \textit{T} denotes time):

\begin{equation}
\label{eq:interface-preference}
{P_{interface}}
= \frac{T_{on\ the\ contact\ area}}{T_{session}}
\end{equation}

Interface preference \eqref{eq:interface-preference} was defined as the chick’s allocation of time to the soft robotic interface area, serving as an indicator of overall acceptance. Its theoretical chance level was determined by the proportion of the arena floor occupied by the interface. Specifically, this area ratio was 0.074 in Experiments 1a and 1b (Fig. \ref{fig_3}a--b), and 0.148 in Experiment 2 (Fig. \ref{fig_3}c). In Experiment 3, the chance level was 0.019, calculated as the ratio of the triangular area formed by the interface edges and vertices to the total arena floor (Fig. \ref{fig_3}d).

\begin{equation}
\label{eq:face-preference}
{P_{face}}
= \frac{T_{on\ the\ contact\ area\ with\ face}}{T_{on\ the\ contact\ area}}
\end{equation}

\begin{equation}
\label{eq:heating-preferencee}
{P_{heating}}
= \frac{T_{on\ the\ contact\ area\ with\ heating}}{T_{on\ the\ contact\ area}}
\end{equation}

\begin{equation}
\label{eq:breathing-preferencee}
{P_{breathing}}
= \frac{T_{on\ the\ contact\ area\ with\ breathing}}{T_{on\ the\ contact\ area}}
\end{equation}

To evaluate specific stimuli, face preference \eqref{eq:face-preference} was quantified as the proportion of interaction time allocated to the interface with the visual faceplate. Similarly, heating preference \eqref{eq:heating-preferencee} and breathing preference \eqref{eq:breathing-preferencee} were defined by the time allocated to the interface with thermal and motion stimuli, respectively. For the face (Experiment 1b) and heating (Experiments 2 and 3) cues, the chance level was set to 0.5, as each stimulus covered exactly half of the available interface area. The breathing cue also used a 0.5 chance level because only one side of the interface was active at a time (counterbalanced across sessions). Across all metrics, a score at the chance level indicates behavioral neutrality, whereas a score above chance indicates a preference for the respective target.

All analyses were conducted in R/RStudio. To analyze the preferences (interface, face, heating, and breathing preference), we fitted Beta mixed-effects models (\textit{glmmTMB} package). This approach is suitable for proportions (0-1 range), allowing the variance to change near the boundaries. We applied the Smithson–Verkuilen transformation to account for 0 and 1 data. Our model was superior to the standard Beta model, the linear mixed model, and the ordered Beta model based on Akaike Information Criterion. Time (six 5-minute time bins) was included as a within-subject fixed effect, and chick was included as a random effect. Model assumptions were checked using \textit{DHARMa}, with no evidence of problematic dispersion or influential outliers. The overall significance of fixed effects was evaluated using Type III Wald $\chi^2$ tests. To test whether preferences differed from random choice, we computed estimated marginal means using \textit{emmeans}, back-transformed them to the 0-1 scale, and compared them against the neutral baseline. For visualization, the result figures show the raw individual data, whereas all data statistics were based on the fitted models. The neutral value (no-preference) was 0.5 for breathing, face, and heating preferences. The no-preference level for the interface was 0.074 in Experiments 1a and 1b; 0.148 in Experiment 2, and 0.019 in Experiment 3, these figures depending on the area occupied by the interface. Statistical significance was set at $p \leq 0.05$.

\section{Experimental Results}
\subsection{Experiment 1a (horizontal)}
In Experiment 1a (N = 10), the interface preference remained stable over time (Fig. \ref{fig_4}a, solid line; $\chi^2(5) = 8.246, p = 0.143$) and stayed around the chance level (mean\_interface = 0.116, $z = 1.045, p = 0.296$). The breathing preference showed a similar pattern (Fig. \ref{fig_4}b, solid line; $\chi^2(5) = 1.652, p = 0.895$) remaining close to the chance level (mean\_breathing = 0.538, $z = 0.936, p = 0.349$). Overall, these results did not provide evidence that the chicks avoided the interface or the rhythmic breathing stimulus.

\begin{figure*}
\centering
\includegraphics[width=7in]{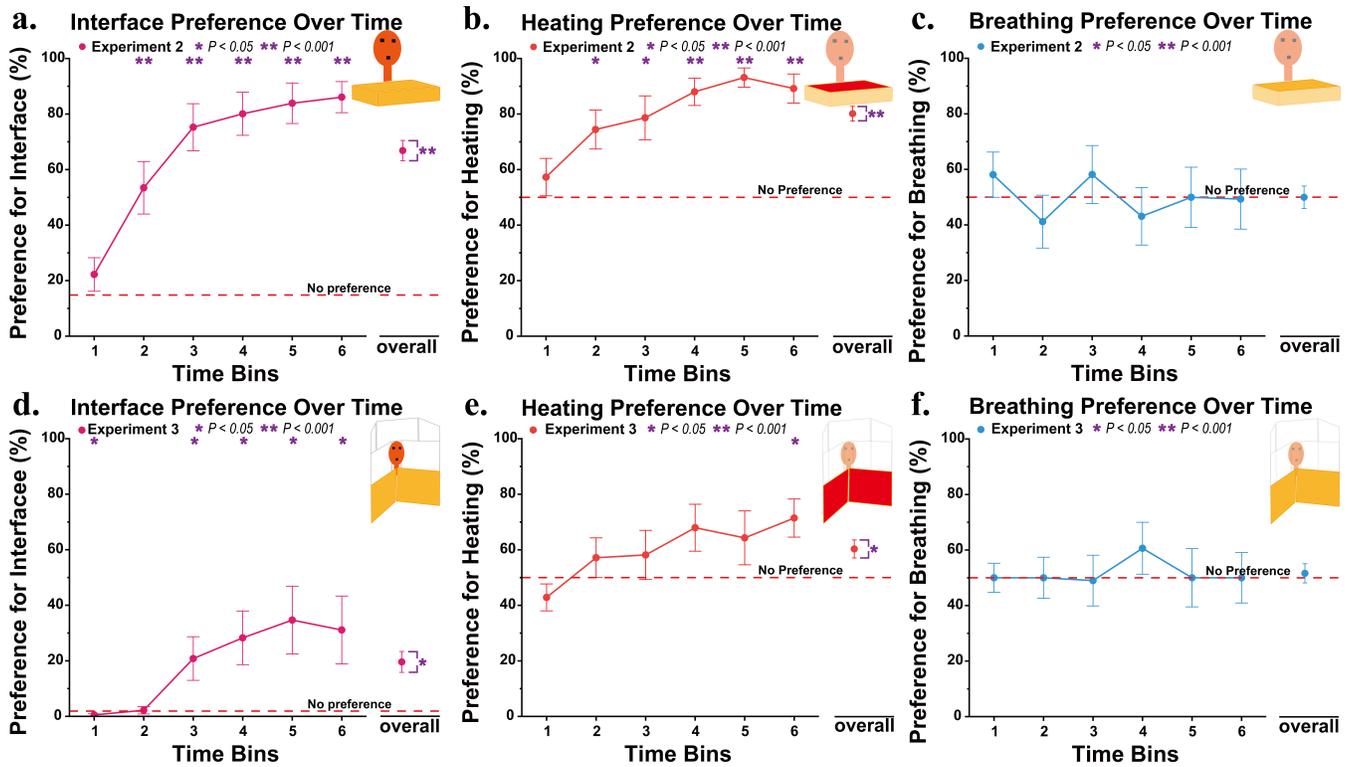}
\hfil
\caption{Preferences in Experiment 2 (horizontal, with faceplate and heating) and Experiment 3 (vertical, with faceplate and heating). Figures display raw data for visualization, whereas all data statistics were based on the fitted Beta mixed-effects models. The x-axis represents the experimental session divided into six consecutive time bins (5 min each). The isolated “overall” data points on the right of each panel indicate the estimated marginal means averaged across all six bins for the entire session. Red dashed lines indicate theoretical chance levels. Error bars represent the standard error of the mean (SEM). Asterisks denote statistical significance (*$p < 0.05$, **$p < 0.001$). \textbf{a.} Interface preference over time in Experiment 2 (horizontal, with faceplate and heating). \textbf{b.} Heating preference over time in Experiment 2. \textbf{c.} Breathing preference over time in Experiment 2. \textbf{d.} Interface preference over time in Experiment 3. \textbf{e.} Heating preference over time in Experiment 3. \textbf{f.} Breathing preference over time in Experiment 3.}
\label{fig_5}
\end{figure*}

\subsection{Experiment 1b (horizontal, with faceplate)}
In Experiment 1b (N = 10), the interface preference increased significantly over time (Fig. \ref{fig_4}a; $\chi^2(5) = 19.226, p = 0.002$). While chicks did not show any significant preference in the first time bin (bin 1: mean = 0.247, $p = 0.070$), they preferred the interface from bin 2 onward (bin 2: mean = 0.371, $p = 0.010$; bin 3--6: mean = 0.512--0.693, all $p < 0.001$). Overall, interface preference was significantly above chance (mean\_interface = 0.529, $z = 4.968, p < 0.001$).
The face preference remained stable across the session (Fig. \ref{fig_4} c; $\chi^2(5) = 6.722, p = 0.242$) and was significantly above chance overall (mean\_face = 0.728, $z = 3.862, p < 0.001$). In contrast, the breathing preference, while also stable across the session (Fig. \ref{fig_4} b; $\chi^2(5) = 4.882, p = 0.430$) remained at chance level (mean\_breathing = 0.496, $z = -0.103, p = 0.918$). Overall, chicks showed a clear preference for the face-like cues, increasing preference for the interface over time, while showing no evidence of preference nor avoidance for the breathing stimulus.

Compared with Experiment 1a, Experiment 1b showed higher acceptance of the apparatus. In Experiment 1a, the interface preference did not change significantly with time ($\chi^2(5) = 8.246, p = 0.143$) and remained close to overall chance (mean\_interface = 0.116; $z = 1.045, p = 0.296$). In Experiment 1b, the interface preference increased across time bins ($\chi^2(5) = 19.226, p = 0.002$) and was significantly above chance values (mean\_interface = 0.529; $z = 4.968, p < 0.001$). The chicks in Experiment 1b also showed a clear overall preference for the face-like cue (mean\_face = 0.728; $z = 3.862, p < 0.001$), and this preference remained stable throughout the session. These results show that adding the face-like cue increased approach to the apparatus and helped maintain the interaction over time. In contrast, the breathing preference remained close to chance in both experiments.

\subsection{Experiment 2 (horizontal, with faceplate and heating)}
In Experiment 2 ($N = 20$), the interface preference increased significantly with time (Fig. \ref{fig_5}a; $\chi^2(5) = 62.163, p < 0.001$). Although the chicks did not exhibit preference in the first time bin (bin 1: mean = 0.201, $p = 0.394$), they preferred the interface from bin 2 onward (bins 2--6: mean = 0.562--0.829, all $p < 0.001$). In general, the interface preference was significantly above chance (mean\_interface = 0.683, $z = 8.619, p < 0.001$). The heating preference also changed significantly over time (Fig. \ref{fig_5}b; $\chi^2(5) = 23.211, p < 0.001$). There was no significant preference for heating in the first time bin (bin 1: mean = 0.589, $p = 0.213$), but this increased from bin 2 onward (bins 2--6: mean = 0.744--0.863, all $p < 0.001$). Overall, the heating cue preference was significantly above chance (mean\_heating = 0.796, $z = 8.966, p < 0.001$). By contrast, the breathing cue preference remained stable across the session (Fig. \ref{fig_5}c; $\chi^2(5) = 1.907, p = 0.862$) at the chance level (mean\_breathing = 0.502, $z = 0.055, p = 0.956$). This experiment indicated that the chicks did not avoid the four horizontal interface arrangements, and spent an increasing amount of time near the interfaces as the sessions progressed, showing a clear preference for heating, though not for breathing, which remained close to chance level.

\subsection{Experiment 3 (vertical, with faceplate and heating)}
In Experiment 3 ($N = 14$), the interface preference changed significantly between time bins (Fig. \ref{fig_5}d; $\chi^2(5) = 12.901, p = 0.024$). The chicks preferred the interface across all time bins, the difference becoming significant as the session progressed (bin 1: mean = 0.128, $p = 0.030$; bin 2: mean = 0.147, $p = 0.019$; bins 3--6: mean = 0.249--0.349, all $p \leq 0.002$). In general, the interface preference was significantly above chance (mean\_interface = 0.245, $z = 4.539, p < 0.001$). The heating preference did not change significantly over time (Fig. \ref{fig_5}e; $\chi^2(5) = 9.740, p = 0.083$), indicating a stable pattern over time, but was significantly above the chance overall (mean\_heating = 0.606, $z = 2.271, p = 0.023$). The breathing preference remained stable (Fig. \ref{fig_5}f; $\chi^2(5) = 1.313, p = 0.934$), close to the chance level (mean\_breathing = 0.517, $z = 0.488, p = 0.626$). These results suggest that the vertical interface arrangement affected the interface contact and weakened the heating effect, whereas breathing remained behaviorally neutral.

\begin{table*}[!t]
\renewcommand{\arraystretch}{1.5}
\caption{Summary of Experimental Results (Transfered)\label{tab:results_summary}}
\centering
\begin{tabular}{|>{\centering\arraybackslash}m{1.2cm}||>{\centering\arraybackslash}m{3.0cm}|>{\centering\arraybackslash}m{4.5cm}|m{5.3cm}|}
\hline
\textbf{Exp.} & \boldmath{$N$} & \textbf{Layout} & \multicolumn{1}{c|}{\textbf{Main Outcomes}} \\
\hline
\textbf{1a} & 10 & Horizontal & Behaviorally neutral and no avoidance.\\
\hline
\textbf{1b} & 10 & Horizontal (with faceplate) & Overall preference for face-like pattern. \newline Overall preference for interface.\\
\hline
\textbf{2} & 20 & Horizontal (with faceplate and heating) & Overall preference for heating. \newline Overall preference for interface.\\
\hline
\textbf{3} & 14 & Vertical (with faceplate and heating) & Overall preference for heating. \newline Overall preference for interface.\\
\hline
\end{tabular}
\end{table*}

\section{Discussion}
A basic requirement for any ARI design is to show that the robotic device is not perceived as threatening and does not trigger avoidance. In Experiment 1a, chicks interacted with the interface at chance level, showing behavioral neutrality without evidence of preference or avoidance. However, in Experiments 1b, 2 and 3, where additional cues were introduced, the interface preference increased across time and remained significantly above chance. Between experiments, chicks were attracted to the interface rather than simply showing no avoidance. This pattern remained stable despite changes in the experimental setup, including the number of interfaces, the size of the face-like visual cue, and the presence of heating.

One factor that contributed to this chick-robot affective interaction was the face-like visual cue. Compared with Experiment 1a, the addition of the faceplate in Experiment 1b produced a clear change in behavior. While chicks in Experiment 1a accepted the interface without showing a preference over the chance, chicks in Experiment 1b showed a significant preference for the faceplate, and increasing interaction with the interface over time. This pattern suggests that the face-like visual cue not only supported acceptance, but also enhanced interaction. This is consistent with previous work showing that face-like visual patterns are highly salient to newly hatched chicks and can guide early orientation and behavior \cite{rosa-salvaFacesAreSpecial2010a}.

Besides the visual cue, heating also played an important role in the interaction. When a heated area was available (Experiments 2 and 3), chicks showed an overall preference for warmth. In Experiment 2, this preference increased over time and was more evident in the later time bins. This finding is consistent with the expectation that newly hatched chicks actively seek thermal comfort \cite{deatonEffectBroodingTemperature1996}. In Experiment 3, heating preference also remained significantly above chance overall, but it did not show a clear increase across time bins. This suggests that the heating effect depended not only on the presence of warmth, but also on the physical layout of the interface, which may have influenced how easily chicks could discover and maintain contact with the heated surface.

Overall, chicks reliably made spontaneous contact with the interface, indicating that the device was generally well accepted. In contrast, the rhythmic breathing cue did not elicit a preference response. This consistent neutrality may reflect several factors. The breathing motion may have been too subtle to elicit any response while still remaining safely non-aversive. Alternatively, the movement pattern or contact area may not have matched the natural experience between a chick and a hen. It is also possible that chicks did not maintain sufficiently sustained full-body contact for the cue to be clearly perceived. It may also reflect the fact that, at this early stage soon after hatching, chicks were more motivated to rest or seek warmth than to respond to more subtle affective cues.

These findings highlight several priorities for ARI development. Future work should move beyond demonstrating basic acceptance and examine how specific stimulus parameters influence behavior within a non-aversive context. Given that the breathing stimulus did not trigger avoidance, we can assume that it provides a safe baseline for further testing. This baseline could be used to explore different actuation intensities, frequencies, and multimodal combinations to identify more effective cues. At the same time, the contact surface and interface geometry should be revised to better match natural chick–hen interaction and support sustained body contact. Behavioral assessment should also extend the time-based measures to include more detailed contact-based measures, such as touch frequency, body part involvement, and postural changes during the contact. Such measures may help clarify which cues are behaviorally meaningful, when they matter, and how they may influence animal welfare. Furthermore, they may support the identification of interaction principles that can be adapted from one species to another.

\section{Conclusion}
This study translates affective interface principles from Human-Robot Interaction to Animal-Robot Interaction. Addressing limited design knowledge for species-specific interactions, we developed an animal-centered, soft robotic interface for newly hatched chicks. Our system combined affective cues: thermal, rhythmic breathing, and face-like visual cues with a repeatable behavioral evaluation protocol based on spontaneous approach and video tracking. Across our experiments, chicks accepted the interface without any evidence of avoidance. We established that visual and thermal stimuli acted as primary drivers of engagement. These cues generated strong preferences that sustained or increased over time, indicating genuine attraction rather than brief novelty effects. Although the rhythmic breathing cue remained behaviorally neutral, it proved non-aversive, establishing a safe baseline for future tactile interaction designs. Ultimately, the main contribution of this paper is the successful design and validation of a multimodal ARI platform, alongside a standardized, welfare-friendly evaluation framework. This work provides a verified technical baseline and clear methodological guidelines for designing species-appropriate robotic interfaces to enhance animal welfare and neuroscientific research.

\section{Acknowledgment}
We thank Mish Toszeghi for his editorial help. We thank Ishani Nanda, Robyn Roach, Antonella Torrisi and staff in the Biological Service Unit (BSU, Queen Mary University of London) for their help during this research.

\bibliographystyle{IEEEtran}
\bibliography{Main}

\vfill

\end{document}